\documentclass[lettersize,journal]{IEEEtran}
\usepackage{algorithmic}
\usepackage{algorithm}
\usepackage{array}
\usepackage[caption=false,font=normalsize,labelfont=sf,textfont=sf]{subfig}
\usepackage{textcomp}
\usepackage{stfloats}
\usepackage{url}
\usepackage{verbatim}
\usepackage{graphicx}
\usepackage{makecell}
\usepackage{cite}
\usepackage{booktabs}
\usepackage{hhline}
\usepackage{hyperref}
\usepackage{multirow}
\usepackage{bbding}
\usepackage{amsthm}
\usepackage{cite}
\usepackage{amsmath,amssymb,amsfonts}
\usepackage{inconsolata}
\usepackage{xcolor}
\usepackage{threeparttable}
\usepackage{framed}
\usepackage{multicol}
\usepackage{mdwlist}
\usepackage{listings}
\usepackage{enumitem}

\usepackage{listings}
\usepackage{xcolor}
\usepackage{xspace}

\newcommand{\tool}{\textsf{LACE}\xspace}
\colorlet{punct}{red!60!black}
\definecolor{background}{HTML}{EEEEE1}
\definecolor{delim}{RGB}{20,105,176}
\colorlet{numb}{magenta!60!black}

\lstdefinelanguage{json}{
    basicstyle=\scriptsize\ttfamily,
    numbers=none,
    numberstyle=\scriptsize,
    stepnumber=1,
    numbersep=8pt,
    showstringspaces=false,
    breaklines=true,
    frame=lines,
    backgroundcolor=\color{background},
    literate=
     *{0}{{{\color{numb}0}}}{1}
      {1}{{{\color{numb}1}}}{1}
      {2}{{{\color{numb}2}}}{1}
      {3}{{{\color{numb}3}}}{1}
      {4}{{{\color{numb}4}}}{1}
      {5}{{{\color{numb}5}}}{1}
      {6}{{{\color{numb}6}}}{1}
      {7}{{{\color{numb}7}}}{1}
      {8}{{{\color{numb}8}}}{1}
      {9}{{{\color{numb}9}}}{1}
      {:}{{{\color{punct}{:}}}}{1}
      {,}{{{\color{punct}{,}}}}{1}
      {\{}{{{\color{delim}{\{}}}}{1}
      {\}}{{{\color{delim}{\}}}}}{1}
      {[}{{{\color{delim}{[}}}}{1}
      {]}{{{\color{delim}{]}}}}{1},
}

\definecolor{BrickRed}{RGB}{178,34,34}

\makeatletter  
\newif\if@restonecol  
\makeatother

\usepackage[
  n,
  operators,
  advantage,
  sets,
  adversary,
  landau,
  probability,
  notions,  
  ff,
  mm,
  primitives,
  events,
  complexity,
  asymptotics,
  keys]{cryptocode}

\begin{document}
\title{``\textit{Say What You Mean}'':\\ Natural Language Access Control with Large Language Models for Internet of Things}

	\author{
 Ye~Cheng, 
 Minghui~Xu,
 Yue~Zhang, 
 Kun~Li,
 Hao~Wu,
 Yechao~Zhang,
 Shaoyong~Guo,
 Wangjie~Qiu,
 Dongxiao~Yu,~\IEEEmembership{Senior Member,~IEEE,} 
 Xiuzhen~Cheng,~\IEEEmembership{Fellow,~IEEE}
 \thanks{Corresponding authors: Minghui Xu and Yue Zhang.}
		\thanks{Y. Cheng, M. Xu, Y. Zhang, K. Li, D. Yu and X. Cheng are with the School of Computer Science and Technology, Shandong University, China (e-mail: \{yech@mail., mhxu@, zyueinfosec@, kunli@, dxyu@, xzcheng@\}sdu.edu.cn).}
        \thanks{H. Wu is with the School of Computer Science, Nanjing University, China (e-mail: hao.wu@nju.edu.cn).}
        \thanks{Y. Zhang is with the School of Cyber Science and Engineering at Huazhong University of Science and Technology, China (e-mail: ycz@hust.edu.cn).}
        \thanks{S. Guo is with the School of Computer Science, Beijing University of Posts and Telecommunications, China (e-mail: syguo@bupt.edu.cn).}
        \thanks{W. Qiu is with the Institute of Artificial Intelligence, Beijing Advanced Innovation Center for Future Blockchain and Privacy Computing, Beihang University, Beijing, China, also with Zhongguancun Laboratory, Beijing, China  (e-mail: wangjieqiu@buaa.edu.cn).}

        \thanks{Copyright (c) 20xx IEEE. Personal use of this material is permitted. However, permission to use this material for any other purposes must be obtained from the IEEE by sending a request to pubs-permissions@ieee.org.}
	}
\IEEEtitleabstractindextext{%
\begin{abstract}
Access control in the Internet of Things (IoT) is becoming increasingly complex, as policies must account for dynamic and contextual factors such as time, location, user behavior, and environmental conditions. However, existing platforms either offer only coarse-grained controls or rely on rigid rule matching, making them ill-suited for semantically rich or ambiguous access scenarios. Moreover, the policy authoring process remains fragmented: domain experts describe requirements in natural language, but developers must manually translate them into code, introducing semantic gaps and potential misconfigurations.
In this work, we present \tool, the \textbf{Language-based Access Control Engine}, a hybrid framework that leverages large language models (LLMs) to bridge the gap between human intent and machine-enforceable logic. \tool combines prompt-guided policy generation, retrieval-augmented reasoning, and formal validation to support expressive, interpretable, and verifiable access control. It enables users to specify policies in natural language, automatically translates them into structured rules, validates semantic correctness, and makes access decisions using a hybrid LLM-rule-based engine. 
We evaluate \tool in smart home environments through extensive experiments. \tool achieves 100\% correctness in verified policy generation and up to 88\% decision accuracy with 0.79 F1-score using DeepSeek-V3, outperforming baselines such as GPT-3.5 and Gemini. The system also demonstrates strong scalability under increasing policy volume and request concurrency. Our results highlight \tool's potential to enable secure, flexible, and user-friendly access control across real-world IoT platforms.

\end{abstract}

}

\maketitle

\IEEEdisplaynontitleabstractindextext
\IEEEpeerreviewmaketitle

\section{Introduction} \label{sec:introduction}

With the rapid advancement of Internet of Things (IoT) technology and the widespread deployment of intelligent devices, IoT platforms have become deeply integrated into critical sectors such as smart homes, industrial control systems, healthcare infrastructures, and urban management. Broadly speaking, modern IoT platforms can be categorized into two types. Consumer-oriented platforms, such as Apple HomeKit~\cite{homekit}, Mi Home~\cite{mihome}, and Samsung SmartThings~\cite{smartthings}, prioritize usability and ease of configuration through trigger-action programming models. However, they typically offer only coarse-grained access control, with role distinctions limited to basic categories like ``owner'' or ``guest.'' In contrast, enterprise-grade platforms such as Alibaba Cloud IoT~\cite{aliiot}, AWS IoT~\cite{awsiot}, and Azure IoT~\cite{azure} provide more sophisticated access control mechanisms, including attribute-based policies and limited forms of context-aware decision-making. In both settings, access control remains a fundamental security layer that determines which users or devices can access which resources, and under what conditions.

However, access control in IoT is no longer a matter of evaluating static roles or enforcing hardcoded rules. As device ecosystems grow increasingly complex and interconnected, access policies must account for a wide range of dynamic contextual factors, including time of day, user behavior, device status, sensor inputs, inferred intent, and even social norms. For instance, consider a scenario where a child in a smart home requests to watch television on a school night. Although the system may be configured to restrict entertainment use after 9 PM, it may also need to consider whether the child has completed homework or whether an emergency situation--such as a fire alarm--is active. These nuanced decisions cannot be reliably handled by rigid rule sets or keyword-based matching, underscoring the need for more intelligent, flexible, and semantically aware access control mechanisms.

Access control in today's IoT systems faces two critical challenges. First, the policy authoring process is fragmented and inefficient. In both consumer and enterprise settings, access policies are typically written in natural language by device owners or security professionals, but must be manually translated into enforceable code by developers. This division of labor creates a semantic gap: non-technical users cannot verify whether the implemented logic reflects their intent, while developers may misinterpret vague or high-level descriptions. As a result, policies are often inconsistent, error-prone, and expensive to maintain. Second, although classical models such as discretionary access control (DAC)\cite{ouaddah2017access}, mandatory access control (MAC)\cite{osborn1997mandatory}, role-based access control (RBAC)\cite{sandhu1998role}, and attribute-based access control (ABAC)\cite{hu2015attribute} provide structured mechanisms for permission assignment, they are largely static and coarse-grained, making them poorly suited to the dynamic, heterogeneous nature of IoT environments. Context-aware access control (CAAC)~\cite{kayes2020survey} attempts to address this by incorporating factors such as time, location, or device status into policy decisions. However, most CAAC implementations rely on rigid keyword matching or predefined logic structures, lacking the semantic understanding needed to generalize across diverse and naturalistic inputs. For example, a policy that allows lights to turn on ``when someone is in the living room'' may not match a request triggered by ``\textit{motion detected by the camera in the living room},'' even though the intent is semantically equivalent. These systems often fail to reason over ambiguous, noisy, or dynamic inputs, leading to incorrect denials or unintended authorizations.

This disconnect between human policy intent and machine-enforceable logic is not merely a usability concern--it poses a fundamental security risk. For instance, a parent may believe they have configured their smart lock to deny guest access after 10 PM, but due to an ambiguous interface or a developer's misinterpretation, the policy may in fact only block access until midnight. As a result, the system silently allows entry at 2 AM, violating the user's original intent without warning or visibility. Such gaps between what users expect and what systems enforce are common in both consumer-grade and enterprise-grade platforms, and are exacerbated by the increasing complexity and context sensitivity of modern IoT deployments.
These observations lead us to a central question:
\textit{Can we design an access control framework that understands user intent expressed in natural language, reasons over dynamic context, and produces verifiable, enforceable decisions?}

To address this, we explore the use of large language models (LLMs) not merely as text generators, but as semantic translators and reasoning engines for access control. LLMs such as GPT-4 exhibit strong capabilities in understanding high-level descriptions, recognizing semantically equivalent expressions, and performing multi-step contextual reasoning--all of which are highly desirable in policy modeling and decision-making. However, LLMs alone are insufficient: their outputs are inherently probabilistic and may introduce hallucinations or inconsistencies, especially in security-sensitive settings.
The core insight of our approach is that LLMs, when properly guided and constrained, can form the foundation of an access control system that is both expressive and trustworthy.

To that end, we propose \tool, the Language-based Access Control Engine, which combines prompt-guided policy generation, retrieval-augmented contextual reasoning, and formal validation techniques to ensure both semantic fidelity and decision correctness.
\tool introduces a hybrid architecture that integrates natural language-driven policy generation and LLM-assisted decision-making with retrieval-augmented reasoning and formal verification. Given a natural language policy description, \tool uses prompt-guided LLMs to synthesize structured access policies in JSON format. These policies are then validated through a multi-stage pipeline that includes natural language inference checks and conflict detection via satisfiability solvers. At runtime, incoming access requests are semantically matched to candidate policies using embedding-based retrieval. For requests involving complex, ambiguous, or context-rich conditions, \tool invokes an LLM to reason over the matched policies and produce an access decision along with an explanation. Finally, Open Policy Agent (OPA) acts as a formal verifier to ensure that the decision complies with the enforced policy set.

We evaluate \tool in smart home environments, where users frequently express access requirements in natural language and context-aware decisions are essential.  
 Our system achieves 100\% verified correctness in access control policy generation across multiple LLMs, following automatic validation of semantics and logic. In access decision tasks, \tool delivers up to 88\% accuracy and 0.79 F1-score using DeepSeek-V3, significantly outperforming baseline models including GPT-3.5 and Gemini. Moreover, \tool demonstrates robust scalability: policy matching remains efficient even with 500+ policies, and response latency is well-managed through hybrid execution and batching. In smart home scenarios, \tool enables users to define and enforce context-rich policies--such as regulating device access based on task completion or emergency events--entirely in natural language. These results highlight \tool's practical viability in enabling secure, interpretable, and intelligent access control across dynamic IoT environments.

In summary, this paper makes the following contributions:
\begin{itemize}
    \item We identify a fundamental gap between human access control intent and formal enforcement, motivating the need for semantic, language-driven policy systems in IoT.
    \item 
We introduce \tool, a hybrid architecture that leverages large language models for policy synthesis and reasoning, while ensuring soundness through structured validation and formal checking.

\item We conduct empirical evaluations across real-world smart home scenarios, demonstrating the effectiveness, transparency, and practical deployability of the approach.
\end{itemize}

\section{Preliminaries}\label{sec: Preliminaries}
\subsection{LLM and Techniques for LLM Inference}

 
LLMs~\cite{zhao2024surveylargelanguagemodels} have emerged as transformative tools in natural language processing, enabled by advances in deep learning and the availability of large-scale training data. They possess several key properties that make them especially effective for language understanding and generation. First, their massive scale, often comprising tens or hundreds of billions of parameters, allows them to capture complex linguistic and semantic patterns and generalize across diverse domains without the need for task-specific retraining. Second, LLMs demonstrate emergent capabilities such as in-context learning and multi-turn reasoning, which are not explicitly programmed but arise during pre-training. These abilities are essential for interpreting high-level or ambiguous natural language inputs. Third, LLMs are highly effective at modeling long-range dependencies and capturing nuanced contextual information, which is critical for understanding inputs influenced by environmental cues, user intent, or temporal and spatial conditions.

While LLMs possess strong general-purpose reasoning abilities, they often struggle with complex, multi-step, or knowledge-intensive tasks when operating without additional guidance. To address these limitations, recent research has introduced techniques such as Chain-of-Thought (CoT) prompting and Retrieval-Augmented Generation (RAG), which significantly improve LLM performance by enhancing their reasoning depth and grounding their outputs in relevant external information: 
\begin{itemize}
    \item \textbf{Chain of Thought Prompting}~\cite{wei2022chain} enhances multi-step reasoning by guiding the model to articulate intermediate steps before arriving at a final answer. This structured reasoning process enables the model to break down complex queries into interpretable elements such as the requesting subject, the target resource, and the associated contextual constraints. By doing so, the model not only improves decision accuracy but also produces transparent and interpretable justifications, which are particularly important in domains where decisions have security or privacy implications.

\item
\textbf{Retrieval Augmented Generation}~\cite{lewis2020retrieval} complements this by improving the factual accuracy and domain alignment of model outputs. Instead of relying solely on internal knowledge acquired during pre-training, this approach incorporates a retrieval component that identifies relevant documents from an external corpus at inference time. The retrieved content is then combined with the input query and passed to the language model to generate a response grounded in real-time evidence~\cite{gao2023retrieval}. This architecture is especially effective for knowledge-intensive applications where factual correctness is critical. In the context of access control, Retrieval Augmented Generation can retrieve candidate policies relevant to a request and guide the model to reason over them to produce or validate a decision.
\end{itemize}

\subsection{Access Control and OPA}

Access control is the process of determining whether a user, service, or application is authorized to perform a specific action on a given resource. In traditional systems, access control policies are often hardcoded and scattered across different services, making them difficult to manage, reuse, and audit. This fragmentation becomes more problematic in modern architectures like microservices and cloud-native environments, where components may rely on inconsistent policy definitions and enforcement mechanisms. 

To address these challenges, the Open Policy Agent (OPA) was introduced as a general-purpose, open-source policy engine that decouples policy from application logic. OPA enables centralized, scalable, and context-aware policy evaluation, allowing for consistent and fine-grained access control across heterogeneous systems.
OPA is a general-purpose, open-source policy engine designed to provide fine-grained, context-aware policy enforcement across heterogeneous systems~\cite{opa}. It decouples decision making from application logic by offering a unified and declarative framework that supports complex authorization use cases in cloud native, distributed, and embedded environments.
OPA expresses policies using Rego, a high-level declarative language rooted in logic programming. Rego allows policies to be specified on structured input data, supporting sophisticated matching, variable binding, and logical inference. At runtime, the OPA evaluates policies by ingesting contextual information and returning decision results through a standard API interface. This design enables external services to offload access control decisions to OPA while maintaining centralized, consistent, and auditable policy logic.

OPA is widely adopted in academic research, such as Kubernetes admission control~\cite{vadisetty2025ai}, web services~\cite{10612535}, identity control~\cite{avirneni2025identitycontrolplaneunifying}, and CI/CD pipelines~\cite{avirneni2025intentawareauthorizationzerotrust}.
In this work, OPA serves as a policy enforcement layer that complements the semantic reasoning capabilities of LLMs. While LLMs provide decisions, OPA acts as a safeguard to enforce hard constraints, validate decisions, and ensure that access control remains interpretable and verifiable.
\section{Design Motivation and Key Idea}

\subsection{Motivation}
Access control in the Internet of Things (IoT) environment is inherently complex. The diversity of devices, the variability in user roles and behaviors, and the dynamic nature of contextual factors such as time, location, and system state all contribute to the challenge. In current practice, access control policies are typically defined using static rule sets that must be manually implemented by developers. These implementations are often based on informal natural language guidance from security experts. This creates a persistent and problematic gap between high-level policy intent and low-level policy code. The process is error-prone, labor-intensive, and difficult to maintain, especially as the number of rules increases and scenarios become more nuanced.

To illustrate the disconnect between human-readable policy intent and machine-executable policy logic, consider the following example: a domain expert specifies the rule ``\textit{Children are allowed to watch television only between 18:00 and 20:00 on weekends}.'' Translating this into a formal policy requires precise identification of the subject, resource, action, effect, and conditional constraints. A corresponding structured policy might be expressed as:
\begin{lstlisting}[language=JSON]
{
  "subject": ["children"],
  "resource": ["television"],
  "action": ["watch"],
  "effect": "allow",
  "conditions": [
    "time >= 18:00",
    "time <= 20:00",
    "day in ['Saturday', 'Sunday']"
  ]
}
\end{lstlisting}

\noindent
While this transformation is straightforward for trivial cases, real-world access control requirements often involve complex, multi-layered conditions and implicit assumptions. As the number and diversity of such policies grow, maintaining consistency, correctness, and interpretability becomes increasingly difficult. Traditional natural language processing (NLP) pipelines, such as those based on syntactic parsing or rule-based extraction, exhibit limited capability in handling linguistic variability, logical dependencies, and contextual nuance. Similarly, conventional policy retrieval mechanisms, based on string-level matching or embedding similarity, are ill-equipped to recognize logical equivalence or contradiction. For example, the statements ``\textit{Children may watch television on weekends}'' and ``\textit{Children may not watch television on weekdays}'' encode equivalent semantics but diverge lexically. Such distinctions are often lost in embedding space representations that prioritize surface-level similarity over formal logical structure.

\subsection{Key Idea}
Large language models have demonstrated impressive capabilities in semantic understanding, contextual reasoning, and natural language generation, making them a promising foundation for rethinking access control in dynamic and heterogeneous environments. Empirical evidence suggests that these models can accurately interpret natural language descriptions of security requirements and translate them into structured representations suitable for policy enforcement. Moreover, they can reason over access requests in context-rich settings, offering human-like explanations for access decisions. However, despite their expressive power, language models are inherently probabilistic. They are prone to hallucinations, semantic drift, and logical inconsistencies--especially in security-critical scenarios where even small errors can lead to policy violations or unintended access. This raises the need for a principled approach that both leverages the strengths of LLMs and compensates for their limitations.

Our core idea is to use LLMs as a semantic interface between natural language policy intent and formal policy enforcement. Instead of relying on traditional rule templates or rigid parsing pipelines, access control policies can be generated through carefully crafted prompts that guide the LLM to extract relevant attributes--such as subjects, resources, actions, effects, and conditions--from natural language descriptions. These generated policies are not directly trusted but undergo semantic verification to ensure they faithfully reflect the original intent. This verification leverages techniques from natural language inference to assess logical equivalence between the generated policy and the expert-provided description.  To ensure logical consistency across multiple policies, each is formalized as a structured tuple, and satisfiability solvers are used to detect conflicts. These include contradictions in effect, redundant rules that add no new semantics, and inconsistencies that result from logically disjoint conditions. This enables the detection and resolution of errors that traditional vector-based similarity models would overlook.
\section{Design of \tool} \label{sec:AC}
 
\begin{figure*}[!htbp]
    \centering
    \centerline{\includegraphics[width=1.0\textwidth, ]{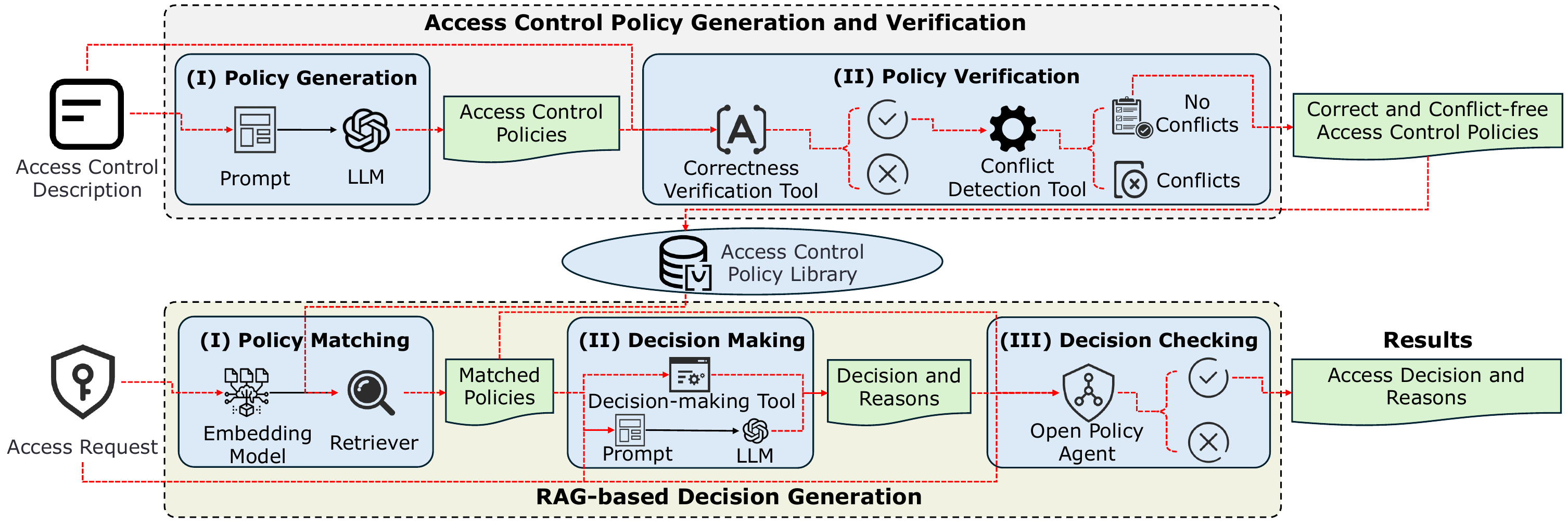}}
	\caption{The architecture of \tool.} 
	\label{fig: architecture}
\end{figure*}

In this paper, we introduce \tool ~(\textbf{L}anguage-based \textbf{A}ccess \textbf{C}ontrol \textbf{E}ngine) a framework designed to enable intelligent, verifiable, and context-aware access control for Internet of Things (IoT) environments. \tool aims to bridge the gap between security experts and system developers by allowing access control policies to be authored in natural language and automatically translated into structured, enforceable rules. The name ``\tool'' reflects the system's core philosophy: \textit{weaving together natural language understanding, formal policy reasoning, and contextual decision-making into a cohesive and secure access control fabric}.  As illustrated in \autoref{fig: architecture}, \tool is composed of two tightly integrated components that collectively support the full lifecycle of access control policy formulation and enforcement.

\begin{enumerate}
\item \textbf{Policy Generation and Verification (\ref{sec:PGV}).} 
This component takes access control descriptions in natural language provided by security experts as input. Firstly, the input is combined with pre-designed prompt templates and fed into an LLM to guide the LLM in generating a series of access control policies in JSON format. Afterwards, these policies undergo correctness verification and conflict detection to ensure that access control policies comply with the descriptions provided by the security experts and that there are no conflicts among them.

\item \textbf{RAG-based Decision Generation (\ref{sec:RDG}).} 
This component takes the user's access request as input and generates the access decision result along with the corresponding rationale. It leverages RAG technology to enhance the accuracy of access control policy matching and optimize the decision-making results of the LLM. First, the request is converted into JSON format and vectorized, based on which matching access control policies are queried from the policy repository. Next, the LLM makes a decision on the access request according to the retrieved policies and outputs the decision result along with the rationale. Finally, it uses OPA to verify the decision result and outputs the final decision. 
\end{enumerate}

\subsection{Policy Generation and Verification} \label{sec:PGV}
\tool employs a two-step process to generate access control policies. First, it utilizes an LLM to generate formatted access control policies based on policy template requirements and natural 
language security specifications. Second, the framework conducts comprehensive correctness and security validation of these candidate policies through formal verification mechanisms. This dual-phase approach ensures the final output satisfies both security expert specifications and conflict-free policy constraints.

\vspace{2mm}
\noindent{\textbf{(Step-I): LLM-based Policy Generation.}}
\tool processes natural language descriptions of access control requirements (e.g.,``\textit{Children are permitted to watch television only between 18:00-20:00 on weekends}.'') to generate formalized access control policies. Leveraging the advanced natural language understanding and inference capabilities of LLMs, it directly synthesizes structured access control policies from textual specifications, effectively handling varied linguistic representations. This methodology replaces traditional NLP tools (e.g., StanfordNLP~\cite{stanford}) by utilizing LLMs' capacity to interpret semantically equivalent formulations across different phrasing conventions. 

\begin{table}[!htbp]  
\centering  
\caption{Prompt $\mathcal{A}$: Access Control Policy Generation}\label{tab:pol_gen}  
\begin{tabular}{p{0.2\columnwidth}p{0.7\columnwidth}} 
\toprule  
\textbf{Elements} & \textbf{Contents} \\  
\midrule  
\textbf{Instruction} & You are a security expert, please convert the natural language described access control description provided by the Input data into formalized access control policies based on the Context. \\ 
\midrule  
\textbf{Context} & 
Basic backgrounds:

1. An Access control policy typically encompasses Subject, Resource, Action, Effect, and Conditions.

2. Subject refers to the entity (e.g., a user or a role) that attempts to perform an action on a resource.

3. Resource represents the object (e.g., files, devices) over which access is being controlled.

4. Action denotes the operations (e.g., view, read, write, control) that a subject wishes to perform on a resource.

5. Effect defines the policy's stance--allowing or prohibiting the action under the conditions specified within the policy itself.

6. Conditions specify the criteria or constraints that must be met for the policy to apply, influencing whether the effect is to permit or deny access.\\  
\midrule  
\textbf{Input data} &  $\langle \mathbf{AC\_description} \rangle$.\\  
\midrule  
\textbf{Output indicator} &  $\langle \mathbf{ACP\_list} \rangle$ in the JSON format:\{``subject'': [``s1'', ``s2'' ...], ``resource'': [``r1'', ``r2'' ...], ``action'': [``a1'', ``a2'' ...], ``effect'': ``allowed or denied'',  ``conditions'': [``c1'', ``c2'' ...]\}. \\  
\bottomrule  
\end{tabular}  
\end{table} 

To ensure the LLM clearly understands the requirements for generating access control policies, we design a prompt (Table~\ref{tab:pol_gen}) that provides specific instructions. This prompt includes background information on the definition and format of access control policies. The input data consists of a list of access control descriptions from users ( $\langle \mathbf{AC\_description} \rangle$). The output, $\langle \mathbf{ACP\_list} \rangle$, is a series of access control policies formatted in JSON.
This approach not only simplifies the process for end-users by allowing them to articulate their security requirements in plain language but also ensures that the generated policies are accurate and tailored to their specifications, thanks to the detailed guidance provided through the prompt.\looseness=-1

\vspace{2mm}
\noindent{\textbf{(Step-II) Policy Verification.}
The generated access control policies $\langle \mathbf{ACP\_list} \rangle$ may have issues related to correctness and security. Correctness means that the meaning represented by the access control policy must be semantically consistent with the original access control description. Security refers to the absence of conflicts between different access control policies.
For example, consider a policy that allows family member A to access the home multimedia devices on Monday, and another policy that prohibits family member A from controlling the home TV on weekdays. Since Monday is part of a weekday, and a TV is considered a multimedia device, these two policies create ambiguity in determining whether family member A should be allowed to access the TV on Monday morning. As a result, these two policies conflict with each other, leading to a situation where a unique decision cannot be made for this specific access request.
Such conflicts can compromise the integrity and reliability of the access control system, making it essential to verify both correctness and security during the policy generation process.

To verify the correctness of access control policies, \tool first reconstructs the policies into sentences in the format of \textbf{Subject} can be \textbf{effect} to \textbf{action} \textbf{resource} if \textbf{conditions}. Subsequently, we employ an advanced AI model specialized in natural language inference (NLI) to perform semantic similarity judgments between the reconstructed sentences and the original access control descriptions. The semantic correctness of access control policies is inherently logic-driven, where even a single-word difference can lead to entirely opposite meanings. For instance, consider the two statements: ``Students are allowed to use their personal phones in the lab on weekends'' and ``Students are not allowed to use their personal phones in the lab on weekends.'' These two sentences differ only by the presence or absence of the word ``not,'' yet their implications are diametrically opposed. When using conventional AI models that rely on vectorization techniques (e.g., embeddings) and compute cosine similarity, the high lexical overlap between these sentences may result in an erroneously high similarity score, failing to capture the critical logical distinction.
In contrast, employing a sophisticated AI model designed for NLI, such as a fine-tuned LLM, can effectively address this limitation. Such models are trained on vast corpora and possess the ability to comprehend nuanced semantic relationships, contextual dependencies, and logical structures within text. By analyzing the broader context and reasoning about the logical implications of negations, conditions, and permissions, these models can accurately distinguish between semantically similar yet logically contradictory statements. This capability ensures that subtle but critical differences in policy formulations are identified, thereby enhancing the precision of access control policy validation.

To verify the security of access control policies, we design a formal modeling and verification approach that systematically detects potential policy conflicts. Specifically, we first model each access control policy as a tuple $$\mathsf{P}=(\mathsf{S}, \mathsf{R}, \mathsf{A}, \mathsf{E}, \mathsf{C}).$$ Here, $\mathsf{S}$ denotes the set of subjects (e.g., users or roles). $\mathsf{R}$ is the set of resources. $\mathsf{R}$ is the set of actions (e.g., view, read, write, control). $\mathsf{E} \in \{Allowed, Denied\}$ is the effect that defines the policy's stance, which allows or prohibits the action under the conditions specified within the policy itself. $\mathsf{C}$ is the set of conditions that specify the criteria or constraints that must be met for the policy to be applied, influencing whether the effect is to permit or deny access. Then, we classify conflicts into three distinct types, consisting of effect conflict, redundancy conflict, and inconsistency conflict. 
\begin{enumerate}
    \item \textbf{Effect Conflict.} An effect conflict occurs when two policies apply to overlapping sets of subjects, resources, and actions under satisfiable, overlapping conditions, yet prescribe contradictory effects--one permitting and the other denying access. Formally, given two access control policies $\mathsf{P_1}=(\mathsf{S_1}, \mathsf{R_1}, \mathsf{A_1}, \mathsf{E_1}, \mathsf{C_1})$ and $\mathsf{P_2}=(\mathsf{S_2}, \mathsf{R_2}, \mathsf{A_2}, \mathsf{E_2}, \mathsf{C_2})$, an effect conflict exists if: $(\mathsf{S_1} \cap \mathsf{S_2} \ne \emptyset) \land (\mathsf{R_1} \cap \mathsf{R_2} \ne \emptyset) \land (\mathsf{A_1} \cap \mathsf{A_2} \ne \emptyset) \land (\mathsf{E_1} \ne \mathsf{E_2}) \land \texttt{SAT}(\mathsf{C_1} \land \mathsf{C_2})$.

    \item \textbf{Redundancy Conflict.} A redundancy conflict arises when one policy is strictly subsumed by another in terms of subject, resource, and action scope, has the same effect, and its condition logically implies the condition of the other. The redundant policy thus contributes no additional access control semantics. This is formally defined as: $(\mathsf{S_1} \subseteq \mathsf{S_2}) \land (\mathsf{R_1} \subseteq \mathsf{R_2}) \land (\mathsf{A_1} \subseteq \mathsf{A_2}) \land (\mathsf{E_1} = \mathsf{E_2}) \land (\mathsf{C_1}\Rightarrow \mathsf{C_2})$.

    \item \textbf{Inconsistency Conflict.} An inconsistency conflict occurs when two policies apply to the same sets of subjects, resources, and actions, and prescribe the same effect, yet their respective conditions are logically disjoint. Although not contradictory in effect, such inconsistency can indicate fragmented or unreachable policy logic. Formally: $(\mathsf{S_1} \cap \mathsf{S_2} \ne \emptyset) \land (\mathsf{R_1} \cap \mathsf{R_2} \ne \emptyset) \land (\mathsf{A_1} \cap \mathsf{A_2} \ne \emptyset) \land (\mathsf{E_1} = \mathsf{E_2}) \land \texttt{UNSAT}(\mathsf{C_1} \land \mathsf{C_2})$.

\end{enumerate}

Next, \tool utilizes SMT solvers (e.g., Z3) to detect conflicts in access control policies based on the formal definitions of these conflict types. Specifically, the SMT solver encodes the policy tuples into logical formulas, enabling systematic analysis of potential conflicts. After correctness verification and conflict detection, \tool outputs the correct and conflict-free access control policies $\langle \mathbf{ACPs} \rangle$ and stores them in the policy repository.

\subsection{RAG-based Decision Generation}\label{sec:RDG}
In traditional access control, decision-making typically relies on static rules or predefined logic. However, these conventional methods may not be flexible enough to handle the complex access request scenarios presented by the IoT. To address this limitation, we propose an access decision-making approach based on RAG, which combines retrieval and generation to achieve smarter and adaptive access control. It is a three-step process. 
First, \tool performs feature extraction and vectorization on the access request, based on which it matches relevant policies from the access control policy repository. Next, it leverages the LLM to make decisions based on the access request and the matched policies. Finally, to ensure the reliability of the decision result, \tool uses OPA to verify the decision. The final output includes the decision result and the rationale behind it.

\vspace{2mm}
\noindent{\textbf{(Step-I): Semantic Policy Matching.}
Traditional retrieval methods in access control systems, such as keyword matching or regular expression filtering, operate on string-level similarity and are unable to capture semantic nuances in natural language or structured attributes. For example, requests such as ``read confidential report'' and ``open restricted document'' may express similar intents but fail to match the same policy due to lexical dissimilarity. 
To address the limitations of traditional policy matching techniques, we propose a semantic matching approach based on the RAG paradigm. We first apply a pretrained sentence embedding model (e.g., Sentence-BERT~\cite{reimers2019sentencebertsentenceembeddingsusing}) to convert each formalized policy $P$ in $\langle \mathbf{ACPs} \rangle$ into a dense vector representation $\texttt{Embed}(P)$. The resulting embeddings are indexed using an approximate nearest neighbor search engine, such as FAISS~\cite{douze2025faisslibrary}. The vector database maintains metadata (e.g., policy IDs) alongside the embedding vectors for efficient retrieval and traceability.

When an access request $\mathsf{Q}=(\mathsf{S}, \mathsf{R}, \mathsf{A}, \gamma)$ is received, where $\mathsf{S}$ is the subject, $\mathsf{R}$ is the resource, $\mathsf{A}$ is the action, and $\gamma$ is contextual information (e.g., time, location, device), we construct a descriptive query string that mirrors the format of policy descriptions. For example, ``User Alice wants to read a confidential file on a mobile device at 9 PM.'' This query is embedded in the same vector space using the same model as policy storage. The embedding $\texttt{Embed}(Q)$ encodes the semantics of the request, including its contextual constraints, ensuring that the retrieval process considers both the core access operation and the situational factors.

Given the embedded request vector $\texttt{Embed}(Q)$, we perform similarity-based retrieval over the policy index. The cosine similarity is used as the distance metric:
$$\texttt{Score}(Q, P_i) = \cos(\texttt{Embed}(Q), \texttt{Embed}(P_i)).$$
The top-$k$ policies $\langle \mathbf{ACP\_matched} \rangle$  with the highest similarity scores are selected as candidates. 

\vspace{2mm}
\noindent{\textbf{(Step-II): Context-aware Decision Making.}
In \tool, access requests and their corresponding policies are first assessed based on their complexity. This strategy is driven by the need to balance computational efficiency with semantic richness in decision-making. While large language models offer powerful reasoning capabilities, invoking them for all requests would introduce unnecessary latency and resource consumption. Not all access decisions require such sophistication; many follow predictable, well-defined rules that can be evaluated deterministically. ``For example, requests such as denying guest access to the front door after 10 PM, preventing children from adjusting the thermostat, or allowing only registered devices to read temperature data typically involve fixed roles, discrete conditions, and clearly bounded resources. '' These scenarios can be processed using traditional rule-based engines with minimal overhead, ensuring fast and reliable responses without sacrificing correctness.

\begin{table*}[!htbp]  
\centering  
\caption{Prompt $\mathcal{B}$: Access Control Decision Making }\label{tab:dec_make}  
\begin{tabular}{p{0.14\textwidth}p{0.8\textwidth}} 
\toprule  
\textbf{Elements} & \textbf{Contents} \\  
\midrule  
\textbf{Instruction} & You are a security expert, please make an access decision based on the access request and the associated access control policies, i.e., whether to allow or deny the request, and provide the reason for the decision. \\ 
\midrule  
\textbf{Context} & 
Basic backgrounds:

1. An Access control policy typically encompasses Subject, Resource, Action, Effect, and Conditions.

2. Subject refers to the entity (e.g., a user or a role) that attempts to perform an action on a resource.

3. Resource represents the object (e.g., files, devices) over which access is being controlled.

4. Action denotes the operations (e.g., view, read, write, control) that a subject wishes to perform on a resource.

5. Effect defines the policy's stance--allowing or prohibiting the action under the conditions specified within the policy itself.

6. Conditions specify the criteria or constraints that must be met for the policy to apply, influencing whether the effect is to permit or deny access.

Follow the  **reasoning steps**  below to determine whether access is permitted:

1. **Verify subject's identity.**

2. **Check each access control policy.**

3. **Analyze fuzzy conditions.** (For example, is the current situation considered an `emergency'? )

4. **Provide a decision (Allow / Deny) along with the rationale.**

5. **Self-check: Review whether the decision complies with all policies. Are there any conflicts?**

6. **If there are errors, correct them and re-evaluate the decision.**\\  
\midrule  
\textbf{Input data} &  $\langle \mathbf{Request} \rangle$, $\langle \mathbf{AC\_matched} \rangle$.\\  
\midrule  
\textbf{Output indicator} &  $\langle \mathbf{AC\_result} \rangle$ in the JSON format:\{``decision'': ``allow or deny'', ``reason'': ``xxx''\}. \\  
\bottomrule  
\end{tabular}  
\end{table*}

In contrast, many real-world access scenarios are inherently ambiguous or highly context-dependent.  
For instance,  consider a more complex scenario in which ``\textit{a child requests to watch television, which is only permitted on weekends after completing homework, and only if no emergency situation (e.g., a fire alarm) is currently active.}'' This request involves multi-layered contextual conditions, including temporal constraints (weekends), task completion status (homework), and dynamic environmental factors (emergency detection). In such cases, 
LLMs offer strong capabilities in semantic understanding, contextual inference, and natural language reasoning, making them well-suited for analyzing access scenarios that go beyond static rule evaluation. By leveraging these capabilities, \tool can interpret complex access semantics, reason over multi-dimensional contextual data, and generate human-understandable rationales for each decision, thereby enhancing both transparency and traceability of the access control process.

To enable LLMs to make such decisions reliably and systematically, we design a dedicated prompt that structures the access decision task as a step-by-step reasoning process. As outlined in Table~\ref{tab:dec_make}, the prompt guides LLMs to follow a chain-of-thought approach, which decomposes complex decision-making into interpretable intermediate steps. The input to the model consists of an access request $\langle \mathbf{Request} \rangle$ and a set of matched access control policies $\langle \mathbf{AC\_matched} \rangle$, while the expected output $\langle \mathbf{AC\_result} \rangle$ includes the final access decision and a textual explanation justifying the outcome. This design enhances the logical consistency of LLM outputs and improves both interpretability and trustworthiness of the access control system.

\vspace{2mm}
\noindent{\textbf{(Step-III): Decision Checking.}
Despite the powerful reasoning capabilities of LLMs, their generative nature may lead to errors in access control decision-making, particularly in scenarios involving subtle conditional logic or conflicting policies. To mitigate these risks and ensure correctness, we integrate an external policy verification module based on OPA. OPA acts as a formal policy compliance engine that verifies the decisions proposed by the LLM against explicitly defined Rego policies. These policies encode the same logical constraints as those used in policy authoring but are compiled into a declarative, verifiable format.

In \tool, every decision output $\langle \mathbf{AC\_result} \rangle$ generated by the LLM is passed to the OPA module along with the original access request and the relevant matched policies. OPA then reevaluates the request using its formal evaluation pipeline and flags any inconsistencies or violations. If the LLM's decision contradicts the outcome derived from OPA, the system either rejects the LLM output or triggers a re-evaluation prompt to guide the model toward a corrected reasoning path. This hybrid decision-making architecture thus combines the flexibility and interpretability of natural language reasoning with the rigor and safety of formal verification.

Furthermore, by logging all mismatches between LLM predictions and OPA validations, we build a feedback dataset that can be used for model fine-tuning or prompt adjustment, enhancing the system's long-term alignment and robustness. In effect, OPA serves not only as a verifier but also as a guardrail that constrains LLM outputs within the bounds of policy soundness and organizational compliance requirements.

\section{Evaluation} ~\label{sec:evaluation}

We have implemented \tool based on the state-of-the-art LLMs, comprising over 2,500 lines of Python code.
To systematically evaluate the effectiveness and efficiency of \tool, we formulate the following research questions:

\begin{itemize}[left=0.8cm]
    \item [\textbf{RQ1}:] \textbf{How accurate and effective is \tool in generating and verifying access control policies?} 
    This question investigates the ability of the system to automatically generate high-quality access control policies and ensure their correctness and security through validation mechanisms.
    \item  [\textbf{RQ2}:] \textbf{How well does \tool perform in RAG-based decision generation, and which LLM provides the best trade-off between accuracy and efficiency?}  
    This question focuses on the decision-making module, where we assess the performance of various LLMs in terms of access decision accuracy and time cost. The goal is to identify the most suitable LLM for practical deployment in access control scenarios.
    \item  [\textbf{RQ3}:] \textbf{How scalable is \tool when handling increasing system loads in terms of the number of policy descriptions, policy entries, and access requests?} This question investigates the scalability of \tool by analyzing the system's performance under varying operational conditions, including the complexity of policy descriptions, the size of the policy library, and the number of concurrent access requests.

    \item [\textbf{RQ4}:] \textbf{How usable and reliable is \tool in enabling natural language-based access control for smart home environments?}
This question examines whether non-technical users can effectively express access control requirements in natural language and whether the system can reliably interpret, enforce, and verify these policies in real-world smart home scenarios. 
\end{itemize}
\subsection{Experiment Setup}~\label{Implementation}

\vspace{2mm}
\noindent{\textbf{Dataset} 
We initially constructed an access control policy dataset consisting of 500 entries, including 300 access control policies sourced from an existing dataset~\cite{jayasundara2024ragent} and 200 additional policies pertaining to IoT devices generated by LLMs. Furthermore, we designed an access decision dataset based on this access control policy dataset. This derived dataset comprises 1000 access requests along with their corresponding access control policies, each manually annotated to indicate whether the request would be allowed or denied. The dataset is utilized to evaluate the performance of the RAG-based decision generation module.

\vspace{2mm}
\noindent{\textbf{Environment}
We selected four local models for policy matching, namely all-MiniLM-L6-v2~\cite{all-MiniLM-L6-v2}, bge-base-en-v1.5~\cite{bge_embedding}, all-distilroberta-v1~\cite{all-distilroberta-v1}, and e5-large-v2\cite{wang2022text}, as well as six popular LLMs for decision making, including Qwen-plus~\cite{qwen-plus}, Qwen-turbo-latest~\cite{qwen-turbo}, Deepseek-V3~\cite{deepseek}, Llama3.3-70B-Instruct~\cite{llama}, GPT-3.5-turbo~\cite{gpt3.5}, and Gemini-2.0-flash~\cite{Gemini2.0}. The performance of these models was evaluated on MacOS Sequoia 15.4.1, Apple M1 Pro CPU, 32GB RAM, and 1TB SSD.\looseness=-1

\subsection{Evaluation Results}~\label{Performance}
\subsubsection{\textbf{RQ1: Effectiveness of Policy Generation and Verification}}
To evaluate the effectiveness of \tool in access control policy generation and verification, we selected four representative LLMs, including Qwen-plus, DeepSeek-V3, GPT-3.5, and Gemini-2.0, to automatically generate policies based on natural language requirements. We assessed the accuracy of generated policies before and after applying the policy verification module. The results are presented in Table~\ref{tab:g_acc}.

\begin{table}[!ht]
\centering
\caption{Accuracy of different LLMs on policy generation}
\label{tab:g_acc}
\begin{tabular}{lcccc}
\toprule
\textbf{LLMs}  & \textbf{Qwen-plus} & \textbf{DeepSeek-v3} & \textbf{GPT-3.5} & \textbf{Gemini-2.0} \\
\midrule
\textbf{Acc(Raw)} & 0.96 & 0.99 & 0.97 & 0.96 \\
\midrule
\textbf{Acc(Final)} & 1.00 & 1.00 & 1.00 & 1.00 \\
\bottomrule
\end{tabular}
\end{table}
As shown in the table, all models achieved high raw accuracy, with DeepSeek-v3 reaching 0.99. After applying the policy verification module, all incorrect or incomplete policies were detected and corrected, resulting in 100\% final accuracy across all models. These results provide strong evidence that \tool can reliably support the end-to-end process of access control policy generation, from natural language specification to semantic validation. More importantly, this outcome aligns with our core motivation: to enable domain users, such as device owners or security experts, to independently express, verify, and reason about access control policies without relying on developers. By eliminating the traditional step of manually translating high-level requirements into code, \tool significantly reduces communication overhead and semantic ambiguity across roles.

\subsubsection{\textbf{RQ2: Accuracy and Efficiency of LLM-Based Decision Making}}
To evaluate the performance of \tool in RAG-based access decision making, we conducted a series of experiments using the access decision dataset focusing on three key aspects: (1) the accuracy of the policy matching component using four embedding models; (2) the decision-making effectiveness of various LLMs under matched policy contexts; and (3) the response latency of LLMs to assess their computational efficiency. These evaluations collectively aim to answer RQ2 by identifying the most suitable LLM for accurate and efficient access control decisions in practical deployments.

Table~\ref{tab:m_acc} shows that for the policy matching step, MiniLM and E5 yield the highest accuracy (both 0.86), making them well-suited for identifying relevant policy candidates from the policy library. Given that the matching step directly influences the input context for LLM-based decision making, selecting a high-accuracy matcher is essential for reliable end-to-end results.
The evaluation results in Table~\ref{tab:llm-evaluation} and Table~\ref{tab:cost} provide insights into the accuracy, efficiency, and overall trade-offs involved in LLM-based access decision making. Among the evaluated models, DeepSeek-v3 achieves the best overall performance in both standalone decision making and end-to-end decision-making scenarios. This demonstrates its robustness in interpreting access requests and reasoning over matched policies.
While Qwen-plus and Qwen-turbo-latest offer slightly lower accuracy, they maintain competitive F1-scores and provide significantly lower inference latency, especially Qwen-turbo-latest, which exhibits the fastest response time of 1.77 seconds per request (Table~\ref{tab:cost}). These results suggest that Qwen-turbo-latest represents a favorable trade-off in scenarios with stricter latency requirements.
GPT-3.5-turbo and Gemini-2.0-flash exhibit considerably lower performance across all metrics, with Gemini particularly underperforming in both recall and overall reliability. Llama3.3-70B-instruct, despite its large size, also underachieves in accuracy and speed, indicating that model size alone does not guarantee improved decision quality in access control contexts.

In summary, DeepSeek-v3 is the most accurate and consistent model for access decision tasks, while Qwen-turbo-latest provides a more efficient alternative with acceptable accuracy. These findings offer practical guidance for deploying \tool in resource-constrained or time-sensitive IoT environments.

\begin{table}[!ht]
\centering
\caption{The accuracy of different models on policy matching}
\label{tab:m_acc}
\begin{tabular}{lcccc}
\toprule
\textbf{Models}  & \textbf{MiniLM} & \textbf{BGE} & \textbf{E5} & \textbf{DistilRoberta} \\
\midrule
\textbf{Accuracy} & 0.86 & 0.79 & 0.86 & 0.70 \\
\bottomrule
\end{tabular}
\end{table}

\begin{table*}[htbp]
\centering
\caption{Evaluation results with different LLMs on decision making task.}
\label{tab:llm-evaluation}
\resizebox{\textwidth}{!}{
\begin{tabular}{l|ccccc|ccccc}
\toprule
\textbf{LLMs} & 
\multicolumn{5}{c|}{\textbf{Decision Making}} & 
\multicolumn{5}{c}{\textbf{End-to-End Decision}} \\

\cline{2-11}
  & \textbf{Accuracy} & \textbf{Precision} & \textbf{Recall} & \textbf{F1-score} &  \textbf{Kappa} &
\textbf{Accuracy} & \textbf{Precision} & \textbf{Recall} & \textbf{F1-score} & \textbf{Kappa} \\

\midrule
Qwen-plus        &  0.86  &  0.66  &  \textbf{0.92}  & 0.77  & 0.67   &  0.74  &  0.51  & \textbf{0.88}  &  0.65 & 0.44\\
Qwen-turbo-latest        &  0.86  &  0.68  &  0.84  &  0.75  &  0.65  &  0.77  &  0.53  &  0.84  &  0.65 & 0.49\\ 
DeepSeek-v3       &  \textbf{0.88}  &  \textbf{0.74}  & 0.85 &  \textbf{0.79}  & \textbf{0.71}  & \textbf{0.86}  &  \textbf{0.70} &  0.85  &  \textbf{0.77} & \textbf{0.67}\\
Llama3.3-70B-instruct  &  0.72  &  0.49  & 0.81 &  0.61  & 0.41  &  0.59  &  0.48  &  0.51  &  0.49 & 0.15\\
GPT-3.5-turbo  &  0.60  &  0.39  & 0.75 &  0.51  & 0.23  &  0.57  &  0.39  &  0.68  &  0.50 & 0.17\\
Gemini-2.0-flash   &  0.59  &  0.59  & 0.29 &  0.39  & 0.13  &  0.57  &  0.54  &  0.29  &  0.37 & 0.09\\

\bottomrule
\end{tabular}
}
\end{table*}

\begin{table*}[t]
\centering
\caption{The time cost of decision making with different LLMs}
\label{tab:cost}
\resizebox{\textwidth}{!}{
\begin{tabular}{lcccccc}
\toprule
\textbf{LLMs}  & \textbf{Qwen-plus} & \textbf{Qwen-turbo-latest} & \textbf{DeepSeek-v3} & \textbf{Llama3.3-70B-instruct}& \textbf{GPT-3.5-turbo}& \textbf{Gemini-2.0-flash} \\
\midrule
\textbf{Time cost} & 3.34s & \textbf{1.77s} & 2.66s & 3.44s & 2.17s& 4.42s\\
\bottomrule
\end{tabular}
}
\end{table*}

\subsubsection{\textbf{RQ3: System Scalability under Varying Load Conditions}}

To evaluate the scalability of \tool, we analyze its time cost performance under three types of increasing system load: the number of access control descriptions used in policy generation, the size of the policy library for matching, and the number of concurrent access requests. These dimensions reflect realistic scalability challenges in dynamic IoT environments.
As shown in Fig.~\ref{fig:combined}(a), the time cost for both policy generation and policy verification grows gradually as the number of access control descriptions increases. Specifically, policy generation time rises from approximately 3.1 seconds (1 description) to 26.4 seconds (100 descriptions), while verification increases more modestly from 0.18 to 0.86 seconds. The generation time exhibits a near-linear trend with respect to the number of descriptions, reflecting the increased prompt size and output length required for policy synthesis.
The primary performance bottleneck in this process lies in the interaction with the LLM itself. As the number of descriptions increases, the input prompt to the LLM becomes significantly longer, and the resulting policy generation also grows in complexity and token count. However, it is worth noting that policy generation is typically a low-frequency operation, occurring primarily during initial system deployment or when policies are updated. As such, latency on the order of several seconds is acceptable and does not impact the real-time performance of the access control system in IoT during normal operation.

In Fig.~\ref{fig:combined}(b), we analyze the effect of increasing the size of the policy library under different levels of concurrent access requests (1, 10, 20). The results indicate that the time cost of policy matching remains nearly constant as the number of policies increases from 50 to 500. For example, when handling 10 requests, the total latency increases only marginally from 0.18 seconds to 0.21 seconds, while for 20 requests, the increase is from 0.34 seconds to 0.38 seconds. This suggests that the scalability of the matching module is primarily affected by the number of concurrent queries rather than the size of the policy repository. 
This efficiency can be attributed to the use of approximate nearest neighbor (ANN) search algorithms in the vector database, which allow for sublinear-time retrieval with respect to corpus size. Such ANN-based retrieval ensures that the system can maintain fast response times even as the policy library grows significantly. Therefore, policy matching is not a bottleneck in \tool and scales well for realistic smart home environments with hundreds or even thousands of policy entries.

Fig.~\ref{fig:combined}(c) illustrates the time cost of \tool under increasing numbers of concurrent access requests, ranging from 1 to 10. To avoid redundant latency caused by repeated communication with LLMs, we adopt a batch processing strategy that groups access requests into a single prompt for inference. This design significantly reduces per-request overhead and improves throughput compared to one-by-one querying. Despite batching, the results show that the LLM-based decision-making component remains the dominant contributor to overall latency. Specifically, decision-making time increases from approximately 3.3s for a single request to over 18.8s for 10 requests, while policy matching and decision checking exhibit much lower and linearly increasing costs. At the maximum load tested (10 requests), decision-making accounts for over 85\% of the total processing time, clearly identifying it as the primary performance bottleneck under high concurrency. Although this is a known limitation of LLMs, it is important to note that in \tool, LLMs are not the default path for all decisions. Traditional rule-based mechanisms are used for standard, deterministic access requests, while LLMs are invoked only when complex, context-rich reasoning is required. In this way, \tool balances the interpretability and flexibility of LLMs with the efficiency of conventional decision logic, ensuring that system-level performance remains acceptable in real-world deployments.

In response to RQ3, our evaluation demonstrates that \tool exhibits strong scalability across key system dimensions. First, the policy generation and verification modules scale efficiently with the number of access control descriptions, with latency gradually increasing and remaining acceptable for low-frequency operations such as initial policy deployment. Second, policy matching remains consistently efficient even as the policy library grows, thanks to the use of ANN retrieval, and its time cost is primarily influenced by the number of concurrent access requests rather than corpus size. Third, while the LLM-based decision-making component incurs the highest latency under load, it is mitigated through batching strategies and system design. LLMs are used selectively only for complex and non-trivial decisions that traditional rule-based mechanisms cannot handle.

\begin{figure*}[!t]
    \centering
    \subfloat[\label{fig:test1}]
    {\includegraphics[width=0.32\textwidth]{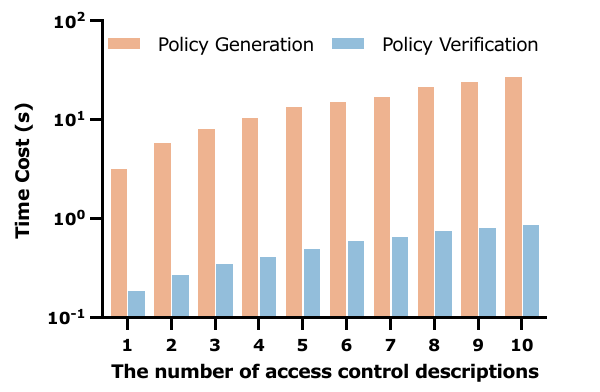}} \hfill
    \subfloat[\label{fig:test2}]
    {\includegraphics[width=0.32\textwidth]{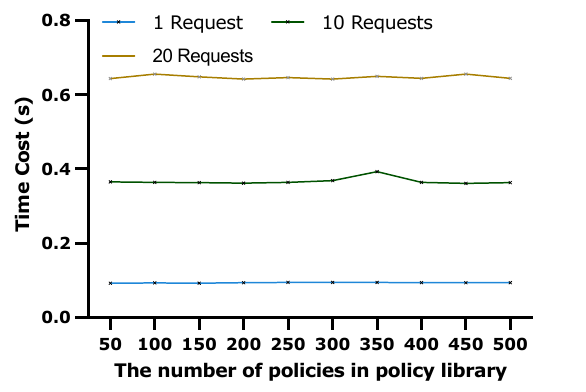}} \hfill
    \subfloat[\label{fig:test3}]
    {\includegraphics[width=0.32\textwidth]{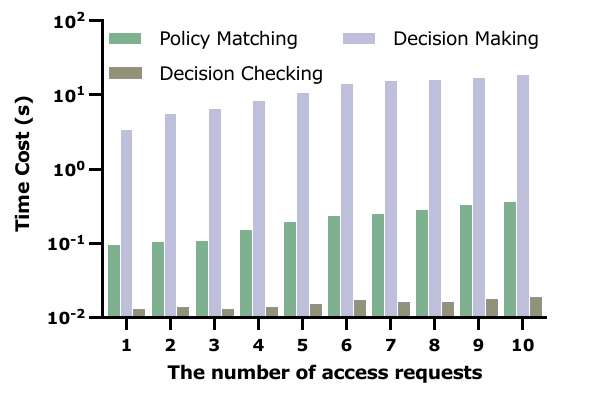}}
    
    \caption{Time cost analysis for different components in \tool under varying loads.}
    \label{fig:combined}
\end{figure*}

\subsubsection{\textbf{RQ4: Demonstrating \tool in Realistic Smart Home Scenarios}}~\label{case study}
We report a case study that employs \tool to assist smart home access control. This case study shows how to integrate \tool into popular IoT platforms. 
In smart home systems, access control policies are often authored through a two-stage process. First, device owners or security experts define the requirements in natural language. Then, developers or system integrators manually implement these policies in code. This separation introduces barriers for non-technical users, who cannot directly express enforceable policies and often struggle to verify whether the final implementation matches their original intent. Even when policies are correctly implemented, the logic is typically opaque to the policy author. As a result, the access control configuration process is inefficient, error-prone, and difficult to audit. To address this, \tool enables users to author and validate access control policies entirely in natural language. There is no need to translate high-level requirements into low-level code, and users can directly review the generated policies for semantic correctness. This significantly lowers the entry barrier for policy creation, while maintaining correctness and transparency in access control enforcement.

In addition, while some smart home platforms support contextual access control, such as enforcing policies based on time, location, or user identity, most systems rely on rigid rule matching based on fixed keywords or values. This limits the system's ability to handle more flexible or ambiguous access conditions. For example, policies involving task completion, inferred intent, or dynamic environments cannot be reliably enforced through traditional means. \tool addresses this limitation by using LLMs to interpret and reason over access requests in a semantically rich manner. Through RAG, the system improves policy matching accuracy, while OPA integration ensures safe execution by verifying decisions before enforcement. In this section, we demonstrate that with \tool, homeowners can not only define expressive access control policies in natural language but also rely on the system to make intelligent, context-aware decisions for complex access requests.

\begin{figure}[!ht]
    \centering
    \centerline{\includegraphics[width=0.49\textwidth]{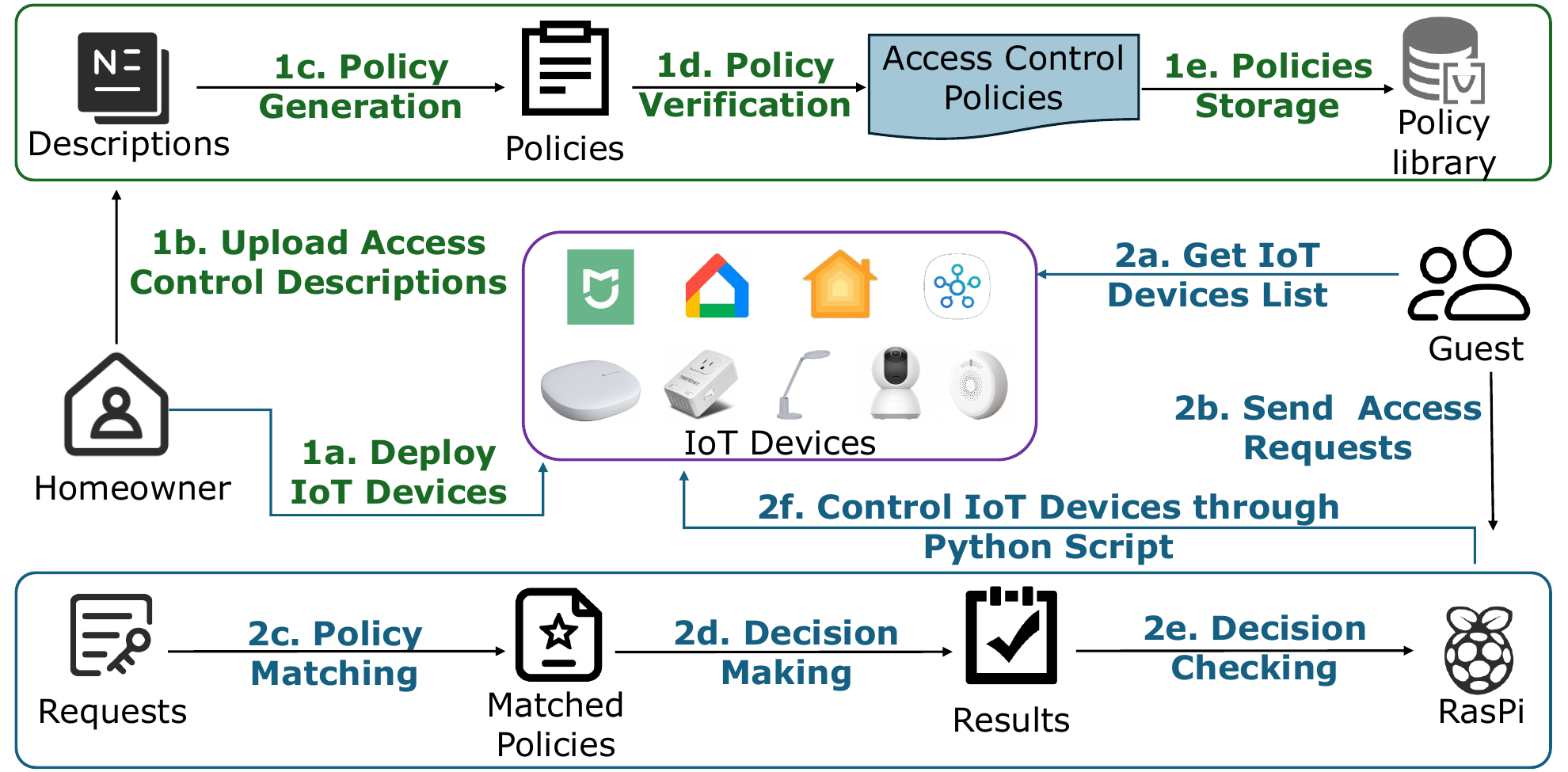}}
	\caption{\tool Assisted Smart Home.} 
	\label{fig: case study}
\end{figure}

Fig.~\ref{fig: case study} shows an \tool assisted smart home case study, divided into two main processes that in that involve a homeowner and a guest. Here's a detailed description following the format provided:
The homeowner's responsibilities commence with the deployment of various IoT devices (1a), which include but are not limited to smart home platforms (e.g., Google Home, Amazon Alexa), smart plugs, lamps, cameras, and sensors. Following the deployment, the administrator uploads access control descriptions (1b) that define the permissible interactions between users and these devices. Based on these descriptions, formalized access control policies are generated (1c). These policies undergo a verification phase (1d) to ensure compliance with established security standards before being stored as vectors in a policy library (1e) for efficient retrieval and application.
For guests seeking interaction with the deployed IoT devices, the process begins with obtaining a list of available devices (2a). Subsequently, guests submit access requests tailored to their needs (2b). The system then matches these requests against the pre-stored policies (2c) to ascertain the legitimacy of the requested access. A decision-making mechanism (2d) evaluates these matches to grant or deny access accordingly. To ensure the decision aligns with both the physical state of devices and real-time constraints, a decision-checking procedure is executed using a Raspberry Pi (RasPi) or similar device (2e). Upon approval, guests can control the IoT devices through Python scripts leveraging libraries such as python-miio (2f) ~\footnote{https://github.com/rytilahti/python-miio}, facilitating automation as per the defined policies. Fig.~\ref{fig: deployment} shows the device deployment topology of our case study. Furthermore, Fig.~\ref{fig: policy} displays a set of some representative access control policies automatically generated by \tool based on natural language descriptions.
\begin{figure}[!ht]
    \centering
    \centerline{\includegraphics[width=0.49\textwidth]{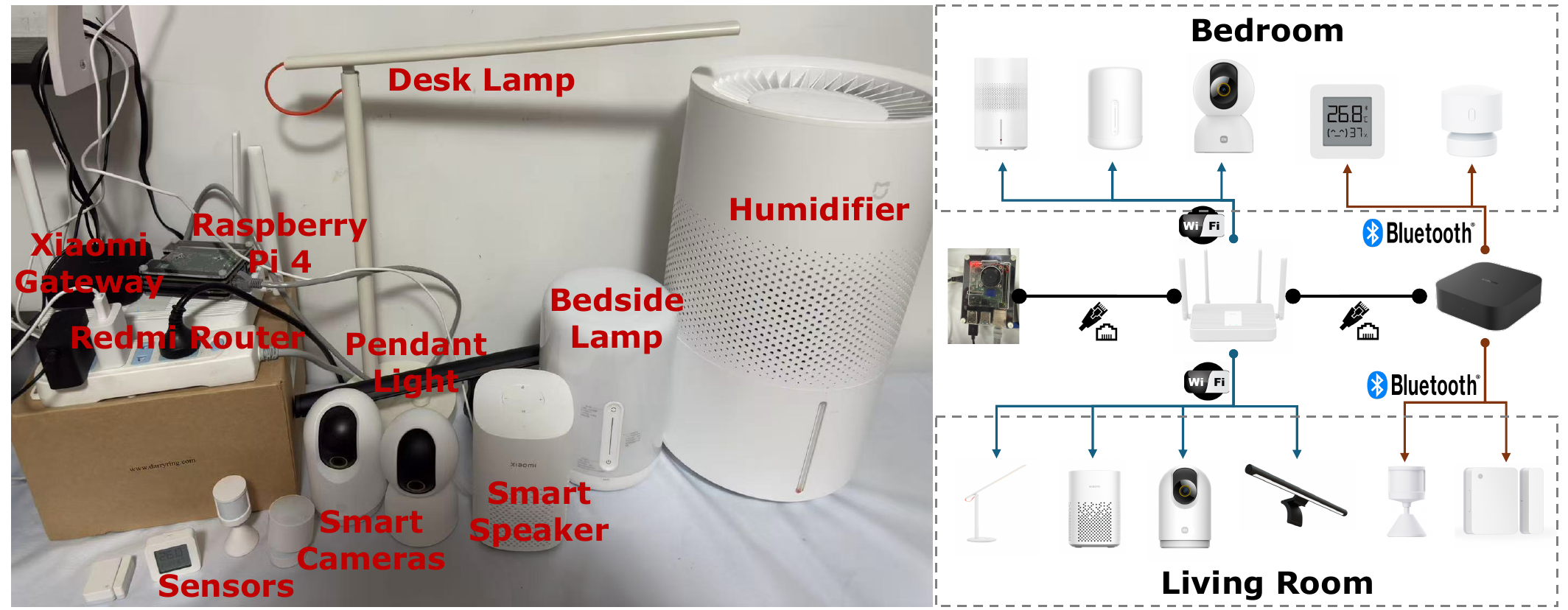}}
	\caption{The device deployment topology of the case study.} 
	\label{fig: deployment}
\end{figure}


\begin{figure}[!htbp]
    \centering
    \begin{lstlisting}[language=JSON]
{
  "policy1": {
    "subject": ["homeowners"],
    "resource": ["smart switches"],
    "action": ["control remotely"],
    "effect": "allowed",
    "conditions": ["authenticate before changes outside of business hours"]
  },
  "policy2": {
    "subject": ["guests"],
    "resource": ["smart plugs"],
    "action": ["operate"],
    "effect": "allowed",
    "conditions": ["within designated time slots approved by the homeowner"]
  },
  "policy3": {
    "subject": ["children under age 16"],
    "resource": ["smart speakers"],
    "action": ["change volume"],
    "effect": "allowed",
    "conditions": ["parental consent between 7 AM and 9 PM"]
  },
  "policy4": {
    "subject": ["all family members"],
    "resource": ["smart air conditioners"],
    "action": ["adjust"],
    "effect": "allowed",
    "conditions": ["significant temperature changes require admin approval"]
  },
  "policy5": {
    "subject": ["visitors"],
    "resource": ["smart doorbells"],
    "action": ["receive temporary access codes"],
    "effect": "allowed",
    "conditions": ["valid only for specific visitation hours"]
  }
\end{lstlisting}
	\caption{Some access control policies generated by \tool.} 
	\label{fig: policy}
\end{figure}

\section{Related Works} 
\label{sec:related work}

In this section, we first summarize the researches on access control in IoT platforms, then conduct an investigation on LLM for access control, and finally describe the limitations of the current access control frameworks in IoT platforms. 
\subsection{Access Control in IoT platforms}
Access control is a foundational mechanism for securing IoT platforms, where diverse users and devices interact across dynamic, often resource-constrained environments. Traditional access control models (e.g., RBAC, ABAC) have been adapted for IoT platforms. 
Sciancalepore~\textit{et al.}~\cite{sciancalepore2016attribute} proposed a streamlined approach for authentication and authorization in IoT platforms using Attribute Based Access Control and token-based techniques. 
Zhang~\textit{et al.}~\cite{8386853} presented a smart contract-based framework for distributed and trustworthy access control in IoT systems, featuring multiple access control contracts, a judge contract, and a register contract to enable both static and dynamic access right validation. 
Sun~\textit{et al.}~\cite{9647918} designed PBAC-FG, a secure and privacy-preserving bilateral access control scheme with fine granularity for cloud-enabled IIoT healthcare, leveraging access control and matchmaking encryption to protect sensitive health data.

However, these models generally assume static roles and permissions, which fall short in dynamic IoT settings that demand fine-grained, context-aware decision making. To address this, recent research has explored context-aware access control frameworks that incorporate environmental factors such as time, location, or device state. 
Arfaoui~\textit{et al.}~\cite{8896965} proposed a context-aware attribute-based access control scheme for IoT that integrates dynamic contextual information with CP-ABE to enable adaptive and efficient data access in highly dynamic environments. 
Merlec~\textit{et al.}~\cite{10454577} presented SC-CAAC, a smart contract-enabled context-aware access control scheme for blockchain-integrated IoT systems, enforcing dynamic and secure access decisions across distributed environments.
Verginadis~\textit{et al.}~\cite{8758387} proposed a holistic context-aware access control framework for cloud computing that extends XACML to enable automated policy federation and infusion into cloud applications.
Dutta~\textit{et al.}~\cite{9123025} designed a context-aware attribute-based access control system for IoT environments, enhanced by intrusion detection for improved smart home security.
While these approaches improve expressiveness, they often require complex manual rule specification and struggle with ambiguity and semantic variation in real-world contexts.

\subsection{LLM for Access Control}
\begin{table*}[t]
\centering
\caption{Comparison of LLM-based Access Control Methods}
\label{tab:llm-ac-comparison}
\resizebox{\textwidth}{!}{
\begin{tabular}{lcccccc}
\toprule
\textbf{Work} & 
\textbf{Policy Generation} & 
\textbf{Correctness Verification} & 
\textbf{Conflict Detection} & 
\textbf{Semantic Policy Matching} & 
\textbf{Context-Aware Decision making} & 
\textbf{Decision Checking} \\
\midrule
IBAC-DB~\cite{subramaniam2024intentbasedaccesscontrolusing} & \Checkmark NL to NLACM  & \Checkmark & \XSolidBrush & \XSolidBrush  & \XSolidBrush & \XSolidBrush \\
LMN~\cite{sonune2025lmntoolgeneratingmachine} & \Checkmark NLACP to MESP & \XSolidBrush & \XSolidBrush & \XSolidBrush & \XSolidBrush & \XSolidBrush \\
Vatsa \textit{et al.}~\cite{vatsa2025synthesizingaccesscontrolpolicies} & \Checkmark NL to JSON & \XSolidBrush & \XSolidBrush & \XSolidBrush & \XSolidBrush & \XSolidBrush \\
iConPAL~\cite{10734044} & \Checkmark NL to policy language & \Checkmark &  \Checkmark & \XSolidBrush & \XSolidBrush & \XSolidBrush\\
\textbf{LACE (Ours)} & \Checkmark NL to JSON & \Checkmark  & \Checkmark & \Checkmark & \Checkmark & \Checkmark \\
\bottomrule
\end{tabular}
}
\end{table*}

Several recent studies have explored the use of LLMs to assist or automate various aspects of access control. Table~\ref{tab:llm-ac-comparison} summarizes recent LLM-based access control frameworks, evaluated along six functional dimensions including natural language policy generation, correctness verification, conflict detection, semantic request-policy matching, context-aware decision making, and decision checking. These criteria reflect key stages in the access control lifecycle and highlight the growing role of LLMs in enhancing automation, adaptability, and semantic alignment.

Some systems, such as LMN~\cite{sonune2025lmntoolgeneratingmachine} and Vatsa et al.~\cite{vatsa2025synthesizingaccesscontrolpolicies}, primarily focus on policy generation from natural language but do not support verification, conflict resolution, or semantic-level decision making. iConPAL~\cite{10734044} and IBAC-DB~\cite{subramaniam2024intentbasedaccesscontrolusing} extend this direction by introducing correctness checking, yet they lack contextual awareness and adaptive decision capabilities.
In contrast, our proposed framework, LACE, offers a comprehensive solution: it supports natural language policy generation, correctness validation (via semantic similarity judgments), conflict detection using satisfiability modulo theories (SMT), RAG-based policy matching, and context-aware decision making, with final checking performed through OPA.

Despite notable advances, existing access control frameworks, both traditional and LLM-based, exhibit several key limitations in the context of dynamic, heterogeneous IoT environments.
First, most context-aware access control models rely on statically defined policies and context descriptors, which require significant manual effort to author and maintain. Such rigidity hinders scalability and fails to accommodate evolving contextual factors or unforeseen usage scenarios.
Second, while LLM-based approaches improve automation and semantic interpretation, they often lack formal guarantees of correctness and consistency. Policies generated by LLMs may appear plausible but diverge from user intent or violate system constraints due to ambiguity, hallucinations, or limited domain grounding.
These limitations highlight the need for a unified, semantically aware, and context-adaptive access control framework capable of maintaining correctness and expressiveness in complex IoT scenarios.
\section{Threats to Validity} \label{sec:validity}

While this work demonstrates the promise of using large language models for access control policy generation and decision-making, several threats to validity remain.

First, our case studies and experiments focus primarily on smart home scenarios, which may not fully represent the diversity of real-world IoT environments such as industrial control systems, healthcare settings, or multi-tenant platforms. The generalizability of our findings to other domains or deployment contexts remains to be validated through broader empirical studies. However, we believe that our modular and domain-agnostic design--particularly the decoupling of natural language interpretation, semantic validation, and policy enforcement--enables \tool to be adapted to new contexts with minimal customization. By separating the policy representation layer from the device-specific execution layer, the system can support varying domains without fundamental changes to its reasoning or verification architecture.

Second, the evaluation of semantic correctness and policy effectiveness is based on both automated checks and human annotations. Natural language interpretation can be inherently subjective, especially in edge cases involving vague expressions or implicit conditions. While we employ carefully designed prompt templates and consistent evaluation guidelines, variation in interpretation may influence both model outputs and human judgments. To mitigate this, we incorporate a multi-stage validation pipeline that combines natural language inference, conflict detection using formal logic, and runtime decision verification through OPA. This layered architecture reduces dependence on any single evaluation method and improves the robustness of our results.

Third, the decision-making process, particularly for complex and context-rich access requests, introduces noticeable latency due to LLM inference and vector-based policy retrieval. This may impact system responsiveness in time-critical environments. For instance, determining whether a caregiver should gain emergency access to a home camera involves multi-layered conditions such as user role, time of day, environmental triggers, and prior approvals, which require contextual interpretation beyond simple rule evaluation. While such decisions are slower to compute, they are often infrequent and high-stakes, where semantic accuracy and explainability are critical. To mitigate this limitation, \tool adopts a hybrid architecture: it uses traditional rule-based engines for routine or low-complexity requests (e.g., ``guests cannot unlock the door after 10 PM'') and reserves LLM-based reasoning for scenarios requiring contextual awareness. Moreover, developers can configure the system based on application requirements, adjusting thresholds for LLM invocation, selecting lightweight models for faster inference, or pre-caching common decision paths to reduce overhead. This flexibility allows \tool to be tailored for performance-sensitive deployments without sacrificing security or correctness in complex scenarios.

\section{Conclusion} ~\label{sec:conclusion}

This paper introduced \tool,  a novel framework that leverages large language models to bridge the gap between natural language policy descriptions and enforceable, verifiable access decisions. By combining prompt-guided policy generation, retrieval-augmented reasoning, and formal verification, \tool empowers users to express and enforce complex access control logic entirely in natural language--without writing a single line of code.  Our evaluation across smart home scenarios demonstrates that \tool achieves high policy correctness, robust decision accuracy, and practical scalability, outperforming several strong baselines. The system's hybrid architecture balances the semantic flexibility of LLMs with the precision of rule-based enforcement and formal validation, making it well-suited for real-world deployment. We envision a future where securing smart environments is as simple as stating your intent and as reliable as formal logic.

\bibliographystyle{IEEEtran}
\bibliography{references}

    \vspace{-1cm}
    \begin{IEEEbiography}[{\includegraphics[width=1in,height=1.25in,clip,keepaspectratio]{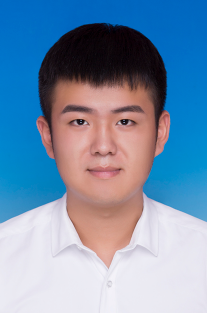}}]{Ye Cheng} received his bachelor's degree in mechanical engineering from Wuhan University of Technology in 2018. He is working toward a Ph.D. degree in School of Computer Science and Technology at Shandong University in China. His current research interests include blockchain and IoT security.
    \end{IEEEbiography}
	 \vspace{-1cm}

	\begin{IEEEbiography}[{\includegraphics[width=1in,height=1.25in,clip,keepaspectratio]{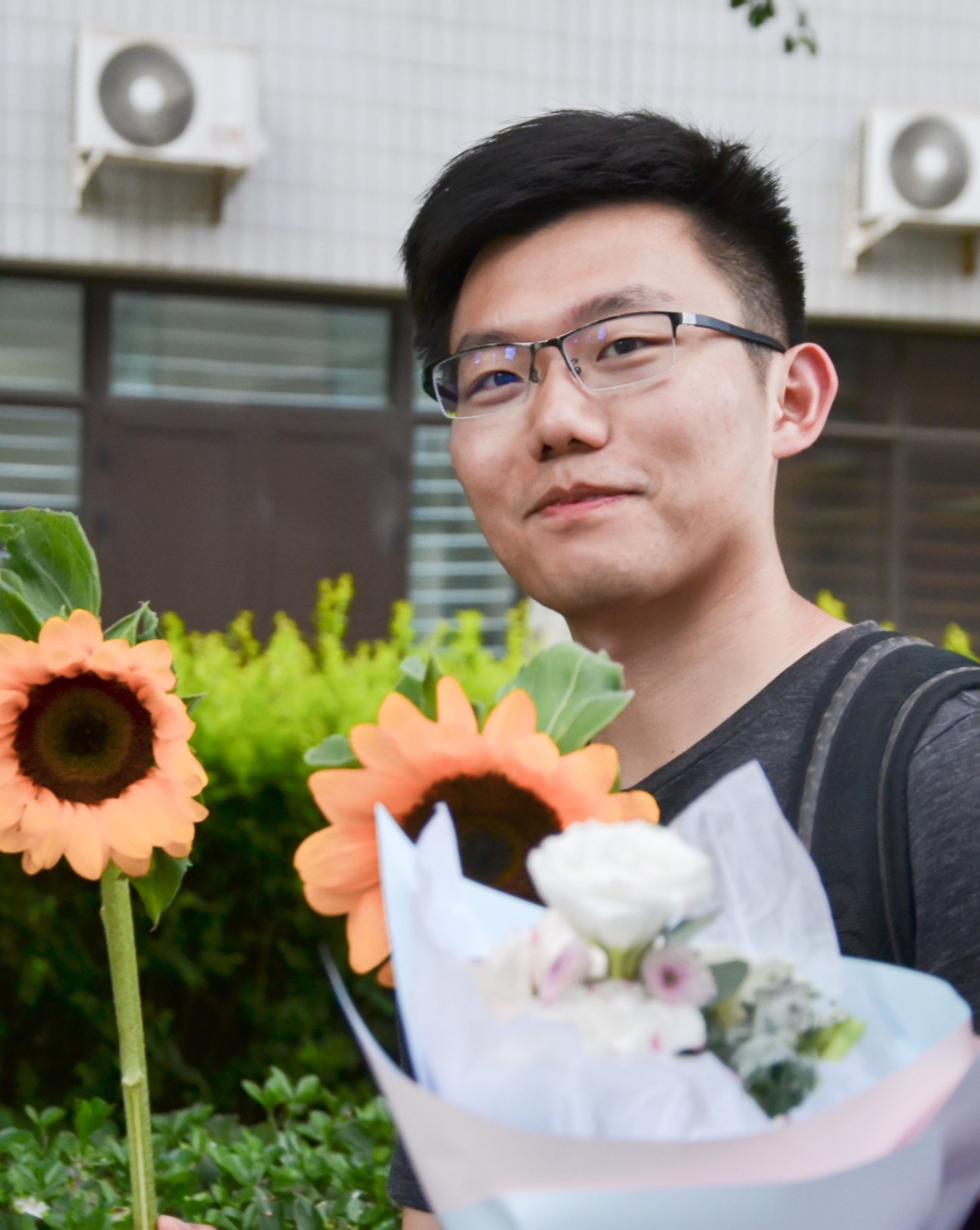}}]{Minghui Xu} is a professor of School of Computer Science and Technology at Shandong University who received his PhD in Computer Science from The George Washington University in 2021 and his Bachelor's degree in Physics from Beijing Normal University in 2018. His research interests include blockchain, distributed computing, and cryptography. 
	\end{IEEEbiography}
	  \vspace{-1cm}

	\begin{IEEEbiography}[{\includegraphics[width=1in,height=1.25in,clip,keepaspectratio]{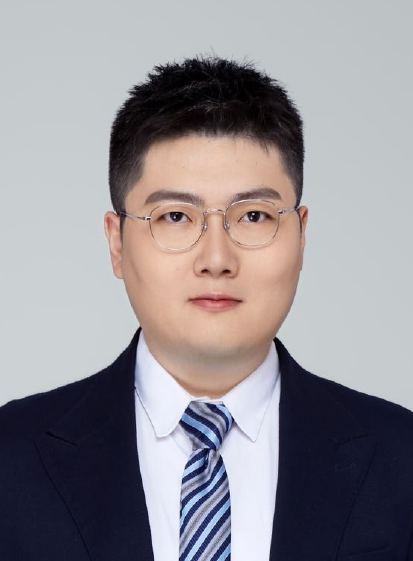}}]{Yue Zhang} is a professor of School of Computer Science and Technology at Shandong University who received the Ph.D. degree from Jinan University in 2020. His research primarily focuses on system security, specifically in the areas of IoT Security and mobile security.  He has published more than 40 papers in security conferences (e.g., USENIX Security, ACM CCS, and NDSS) and journals (e.g., TDSC, TPDS). He received a Best Paper Honorable Mention Award at ACM CCS 2022, and the Best Paper Award at 2019 IEEE International Conference on Industrial Internet. He has also served on the organization committees of the conferences (e.g., general chair of EAI ICECI, track chair for IEEE MSN and IEEE MASS) and technical program committee of the conferences (e.g., USENIX Security, NDSS, ACM CCS, RAID). He serves as an Associate Editor for IEEE T-IFS, HCC and Editor Member of the Blockchain Journal, Electronics Journal, and CMC. His research had led to the discovery of many vendor-acknowledged vulnerabilities, such as by Bluetooth SIG, Apple, Google, and Texas Instruments, and had attracted intense media attention such as Hacker News, and Mirage News. 
	\end{IEEEbiography}
	  \vspace{-1cm}
       
	\begin{IEEEbiography}[{\includegraphics[width=1in,height=1.25in,clip,keepaspectratio]{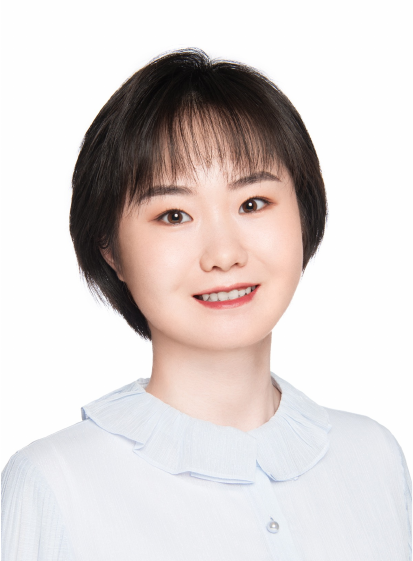}}]{Kun Li} received the Ph. D. degree in 2023 from School of Artificial Intelligence, Beijing Normal University. Now, she is an Assistant Professor at Shandong University. Her research interests include data elements, mobile computing, and blockchain. 
	\end{IEEEbiography}
	  \vspace{-1cm}

	\begin{IEEEbiography}[{\includegraphics[width=1in,height=1.25in,clip,keepaspectratio]{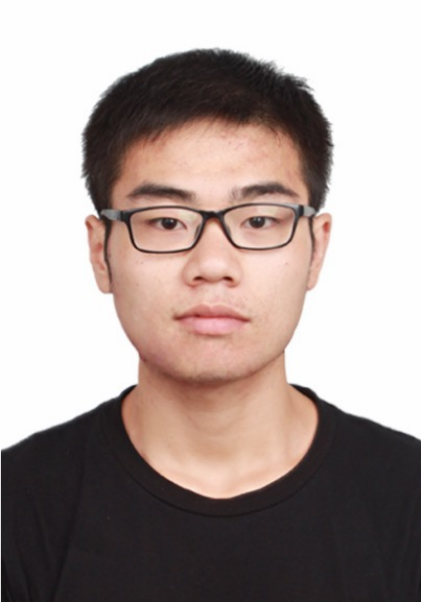}}]{Hao Wu} received his Ph.D. degree from Nanjing University in 2021. He is currently an Assistant Professor at the School of Computer Science, Nanjing University. His dissertation was awarded the Distinguished Dissertation Award by the Chinese Information Processing Society of China, the Jiangsu Computer Society of China, and ACM SIGBED China. His research interests include intelligent mobile computing and AI security.

	\end{IEEEbiography}
         \vspace{-1cm}

	\begin{IEEEbiography}[{\includegraphics[width=1in,height=1.25in,clip,keepaspectratio]{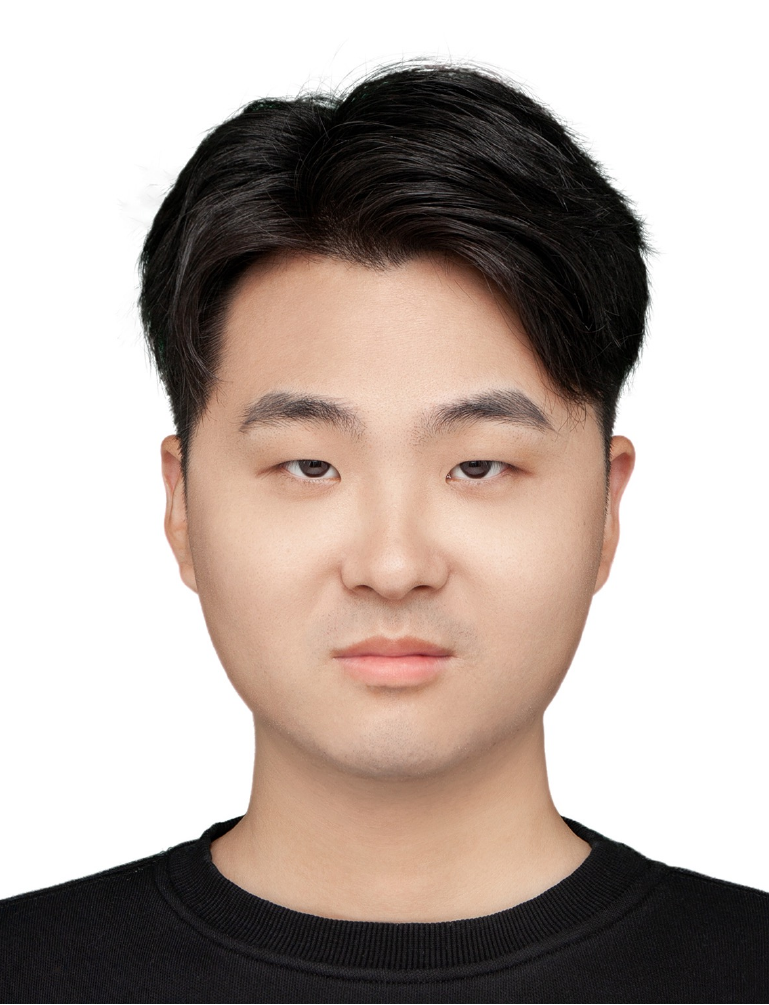}}]{Yechao Zhang} is currently a PhD candidate in the School of Cyber Science and Engineering at Huazhong University of Science and Technology (HUST) in China. He has presented papers at conferences focused on security and AI, including IEEE S\&P, NeurIPS, ESORICS, AAAI, and ACM MM. His research interests include AI security and the application of LLMs for security purposes.

	\end{IEEEbiography}
         \vspace{-1cm}

	\begin{IEEEbiography}[{\includegraphics[width=1in,height=1.25in,clip,keepaspectratio]{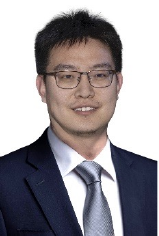}}]{Shaoyong Guo} is a professor at the School of Computer Science, Beijing University of Posts and Telecommunications. He received the National Science Fund for Excellent Young Scholars in 2023. His research interests include Management and Control for Industrial Internet Network, and so on. He is undertaking many key research and development projects and fund projects, and contributed to a number of pioneering standards proposals in ITU-T. The systems and devices developed by him have large-scale application. He was awarded the first prize of Science and Technology Progress Award of Jiangsu Province, and the second prize of Science and Technology Progress Award of Chinese Institute of Electronics, and so on. 
	\end{IEEEbiography}
	  \vspace{-1cm}

	\begin{IEEEbiography}[{\includegraphics[width=1in,height=1.25in,clip,keepaspectratio]{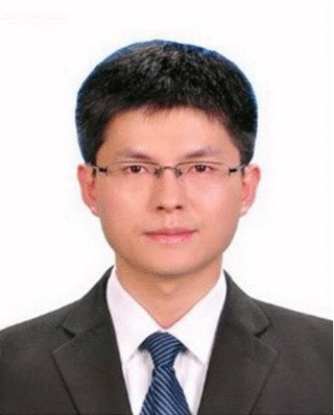}}]{Wangjie Qiu} received the B.S. degree, in 2008, and the Ph.D. degree, in 2013, both in mathematics from Beihang University. He is currently an Associate Professor with Beijing Advanced Innovation Center for Future Blockchain and Privacy Computing, Beihang University. He is also the Deputy Secretary General of the blockchain committee, China Institute of Communications. His research interests include Information security, blockchain, and privacy computing. As the founding team, he successfully developed ChainMaker, an international advanced blockchain technology system. He has been authorized a number of invention patents in the fields of information security, blockchain, and privacy computing. He won the first prize of China National Technology Invention in 2014.
	\end{IEEEbiography}
         \vspace{-1cm}

        \begin{IEEEbiography}
        [{\includegraphics[width=1in,height=1.25in,clip,keepaspectratio]{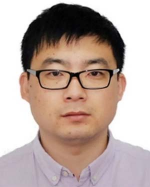}}]{Dongxiao Yu} received the B.Sc. degree in mathematics from the School of Mathematics, Shandong University, Jinan, China, in 2006, and the Ph.D. degree in computer science from the Department of Computer Science, the University of Hong Kong, Hong Kong, in 2014. In 2016, he became an Associate Professor with the School of Computer Science and Technology, Huazhong University of Science and Technology, Wuhan, China. He is currently a Professor with the School of Computer Science and Technology, Shandong University. His research interests include wireless networks, distributed computing, and graph algorithms.
	\end{IEEEbiography}
	  \vspace{-1cm}
  	\begin{IEEEbiography}
        [{\includegraphics[width=1in,height=1.25in,clip,keepaspectratio]{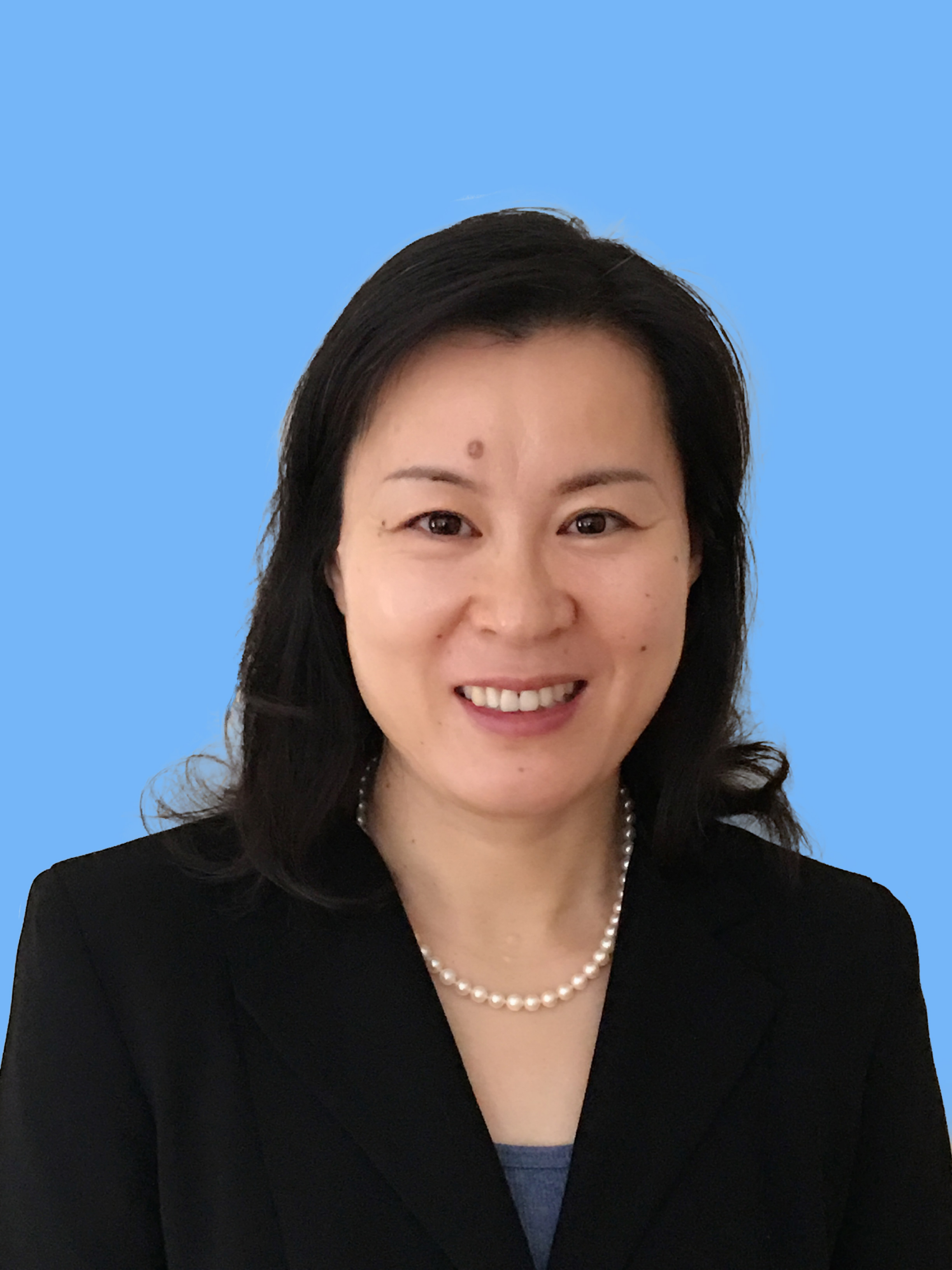}}]{Xiuzhen Cheng} received her MS and PhD degrees in computer science from University of Minnesota, Twin Cities, in 2000 and 2002, respectively. She was a faculty member at the Department of Computer Science, George Washington University,  from 2002-2020. Currently she is a professor of computer science at Shandong University, Qingdao, China. Her research focuses on blockchain computing, security and privacy, and Internet of Things. She is a Fellow of IEEE, a Fellow of CSEE, and a Fellow of AAIA. 
	\end{IEEEbiography}

\end{document}